\documentclass[10pt,twocolumn,letterpaper]{article}

\usepackage[pagenumbers]{cvpr} 
\usepackage{graphicx}
\usepackage{amsmath}
\usepackage{amssymb}
\usepackage{booktabs}
\usepackage{color}
\usepackage{colortbl}  
\usepackage{xcolor}
\usepackage{array}
\usepackage{bbding}
\definecolor{hollywoodcerise}{rgb}{0.96, 0.0, 0.63}
\definecolor{lasallegreen}{rgb}{0.03, 0.47, 0.19}
\definecolor{hanpurple}{rgb}{0.32, 0.09, 0.98}
\definecolor{green(pigment)}{rgb}{0.0, 0.65, 0.31}
\usepackage{multirow}
\newcommand{\aka}{\textit{a.k.a.}}
\usepackage{listings}

\usepackage{xcolor}
\definecolor{mygreen}{rgb}{0,0.6,0}
\definecolor{mygray}{rgb}{0.5,0.5,0.5}
\definecolor{mymauve}{rgb}{0.58,0,0.82}
\usepackage[lined,boxed,commentsnumbered,ruled]{algorithm2e}

\definecolor{cvprblue}{rgb}{0.21,0.49,0.74}
\usepackage[pagebackref,breaklinks,colorlinks,citecolor=cvprblue]{hyperref}

\def\paperID{6652} 
\def\confName{CVPR}
\def\confYear{2024}

\title{OmniBind: Teach to Build Unequal-Scale Modality Interaction for \\ Omni-Bind of All
}

\begin{document}

\author{Yuanhuiyi Lyu$^{1}$ \quad Xu Zheng$^{1}$ \quad Dahun Kim$^{3}$ \quad Lin Wang$^{1}$$^{,2}$ \thanks{Corresponding author.}\\
$^{1}$ VLIS LAB, AI Thrust, HKUST(GZ) \quad $^{2}$Dept. of CSE, HKUST \quad $^{3}$ Google DeepMind
\\
{\tt\small yuanhuiyilv@hkust-gz.edu.cn, zhengxu128@gmail.com, mcahny@google.com, linwang@ust.hk} \\
\small{Project Page: \url{https://vlislab22.github.io/OmniBind/}}
}

\twocolumn[{
\renewcommand\twocolumn[1][]{#1}%
\maketitle
\begin{center}
\centering
\vspace{-20pt}
  \includegraphics[width=\textwidth]{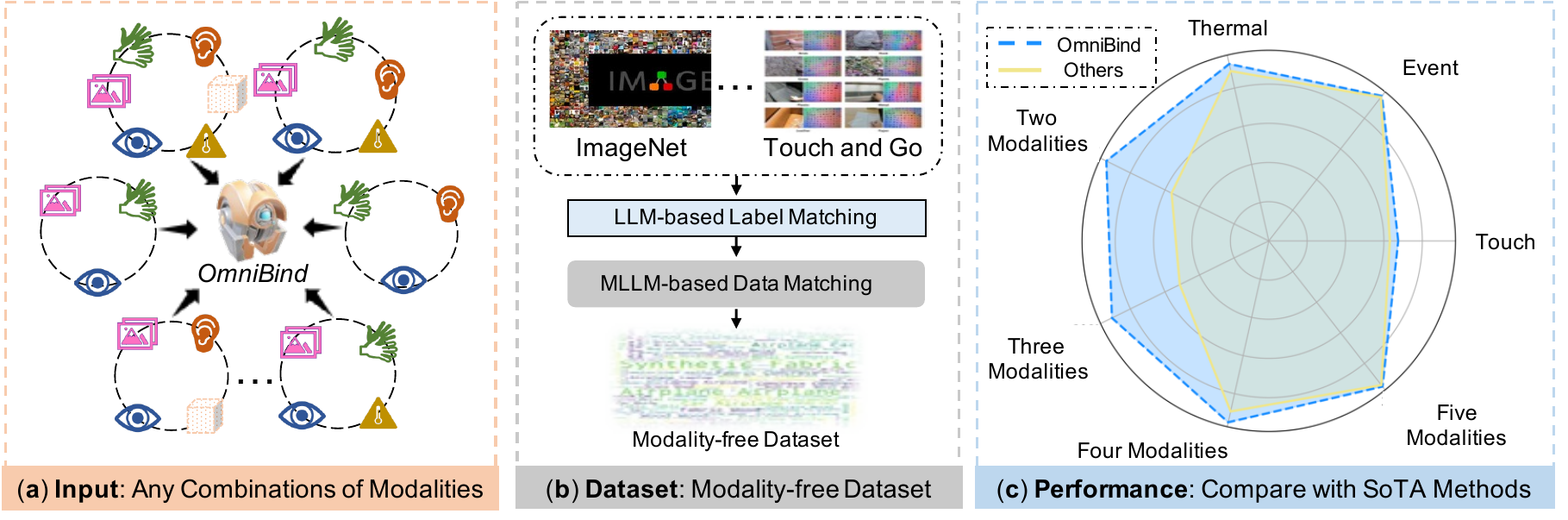}
\captionsetup{font=small}
\end{center}
\vspace{-10pt}
\captionof{figure}{The proposed OmniBind: a novel two-stage learning framework that can achieve any modality combinations and interaction. \textbf{(a)}: The ability that accepts input of any modality combinations. \textbf{(b)}: The performance of our OmniBind. \textbf{(c)}: Modality-free dataset.}
\vspace{15pt}
\label{fig:teaser}
}]

\maketitle
\renewcommand{\thefootnote}{}
\footnotetext{$^*$Corresponding author.}
\begin{abstract}

Research on multi-modal learning dominantly aligns the modalities in a unified space at training, and only a single one is taken for prediction at inference. However, for a real machine, \eg, a robot,  sensors could be added or removed at any time. Thus, it is crucial to enable the machine to tackle the mismatch and unequal-scale (or data-scale imbalance) problems of modality combinations between training and inference. In this paper, we tackle these problems from a new perspective: "\textit{Modalities Help Modalities}". 
Intuitively, we present \textbf{OmniBind}, a \textbf{\textit{novel}} two-stage learning framework that can achieve any modality combinations and interaction. 
It involves teaching data-constrained, \aka, \textit{student}, modalities (\eg, touch and thermal) to be aligned with the well-trained data-abundant, \aka, \textit{teacher}, modalities (\eg, image and text). This subtly enables the adaptive fusion of any modalities to build a unified representation space for any combinations. 
Specifically, 
we propose Cross-modal Alignment Distillation (\textbf{CAD})
to address the unequal-scale problem between student and teacher modalities and effectively align student modalities into the teacher modalities' representation space in stage one.
 We then propose an Adaptive Fusion (\textbf{AF}) module to fuse any modality combinations and learn a unified representation space in stage two. 
To address the mismatch problem, we aggregate existing datasets and combine samples from different modalities by the same semantics. This way, we build the \textbf{\textit{first}} dataset for training and evaluation that consists of teacher (image, text) and student (touch, thermal, event, point cloud, audio) modalities and enables omni-bind for any of them, \eg, (image, touch, and event). 
Extensive experiments show performance gains over prior arts by an average of \textit{\textbf{4.05\%}} on the arbitrary modality combination setting. It also achieves the state-of-the-art performance for a single modality, \eg, touch, with \textit{\textbf{4.34\%}} gain. 

\end{abstract}
\section{Introduction}


Humans are naturally equipped with multiple sensing modalities~\cite{moreno2007interactive, ngiam2011multimodal}, such as vision, audio, and touch, to understand and perceive the world around us. 
This has inspired many recent research endeavors for multi-modal foundational models, which can be divided into two categories: 1) multi-encoder-based and 2) universal-transformer-based methods.
The first line of methods utilizes a specific modality, \eg, image~\cite{girdhar2023imagebind, guo2023point, yang2024binding} or language~\cite{zhu2023languagebind,lyu2024unibind}, as the alignment center for aligning all modalities to the centric modality.
The universal transformer methods~\cite{zhang2024learning,lei2023vit,zhang2023meta} utilize a unified transformer for all modalities via modality-specific tokenizers and prompts.

However, in a real-world machine, such as a robot designed to imitate human perception, the configuration of sensors is not static. Sensors may be added, removed, or replaced at any time~\cite{yu2022all, ilami2021materials} due to various factors such as environmental conditions, specific task requirements, or technological failures~\cite{iba2005interactive,xue2020progress}. Thus there exists a mismatch between training and inference if all modalities are used for training and any combinations of these modalities are expected to be achieved during inference.
Besides, there is still a significant data-scale imbalance between different modalities.
For example, the large gap in data-scale between image modality (\ie, 400 million image data samples) and touch modality (\ie, 12 k data samples). 

Therefore, it is essential to equip the machine with the capability to address the mismatch and data-scale imbalance problems of modality combinations between training and inference stages.
As depicted in Fig.~\ref{fig:teaser} (a), a machine should possess the capability to integrate any modality combinations at any time, such as image, audio, and point cloud, to achieve robust and reliable scene perception. This flexibility would enable the machine to adapt to different sensor configurations and continue to perform accurately.
In light of this, we introduce \textbf{OmniBind}, a two-stage multi-modal learning framework for interacting with any modalities and learning a unified representation space for any modality combinations. The core insight of OmniBind is "\textit{Modalities Help Modalities,}". It involves teaching data-constrained, \aka, \textit{student}, modalities (\eg, touch and thermal) to be aligned with well-trained data-abundant, \aka, \textit{teacher}, modalities (\eg, image and texts). This subtly enables the adaptive fusion of any modalities to build a unified representation space for any combinations.

To make this possible, we first introduce the \textbf{\textit{Cross-modal Alignment Distillation}} (\textbf{CAD}) (Sec.~\ref{sec:3.1}) to tackle the data-scale imbalance problems. It effectively aligns student modalities to the teacher modalities' representation space at two aspects, \ie, the self-correspondence within each modality and the cross-correspondence between teacher and student modalities.
We then propose an \textit{\textbf{Adaptive Fusion}} (\textbf{AF}) module (Sec.~\ref{sec:3.2}) designed to integrate multi-modal representations from any modality combinations. It facilitates the learning of a unified representation space and consistently demonstrates robust performance across diverse modality combinations.

To address the mismatch problem and train our OmniBind, we introduce a modality-free dataset (Sec.~\ref{sec:3.3}), the first dataset that consists of seven modalities and enables omni-bind for any of them, \eg, (image, touch, and event). As shown in Fig.~\ref{fig:teaser} (b), for aggregating this dataset, we combine the data from different modalities via two-level matching: label-level and sample-level matching.

We evaluate our OmniBind on \textbf{1)} arbitrary modality combination setting and \textbf{2)} single modality setting. As shown in Fig.~\ref{fig:teaser} (c), our OmniBind achieves any modality combinations, including combinations of 2-5 modalities. OmniBind shows the remarkable performance of a \textbf{4.05\%} gain on the arbitrary modality combination setting and also achieves state-of-the-art performance for a single modality, such as a notable \textbf{4.34\%} gain for the touch modality.


\begin{figure*}[t!]
    \centering
    \includegraphics[width=\textwidth]{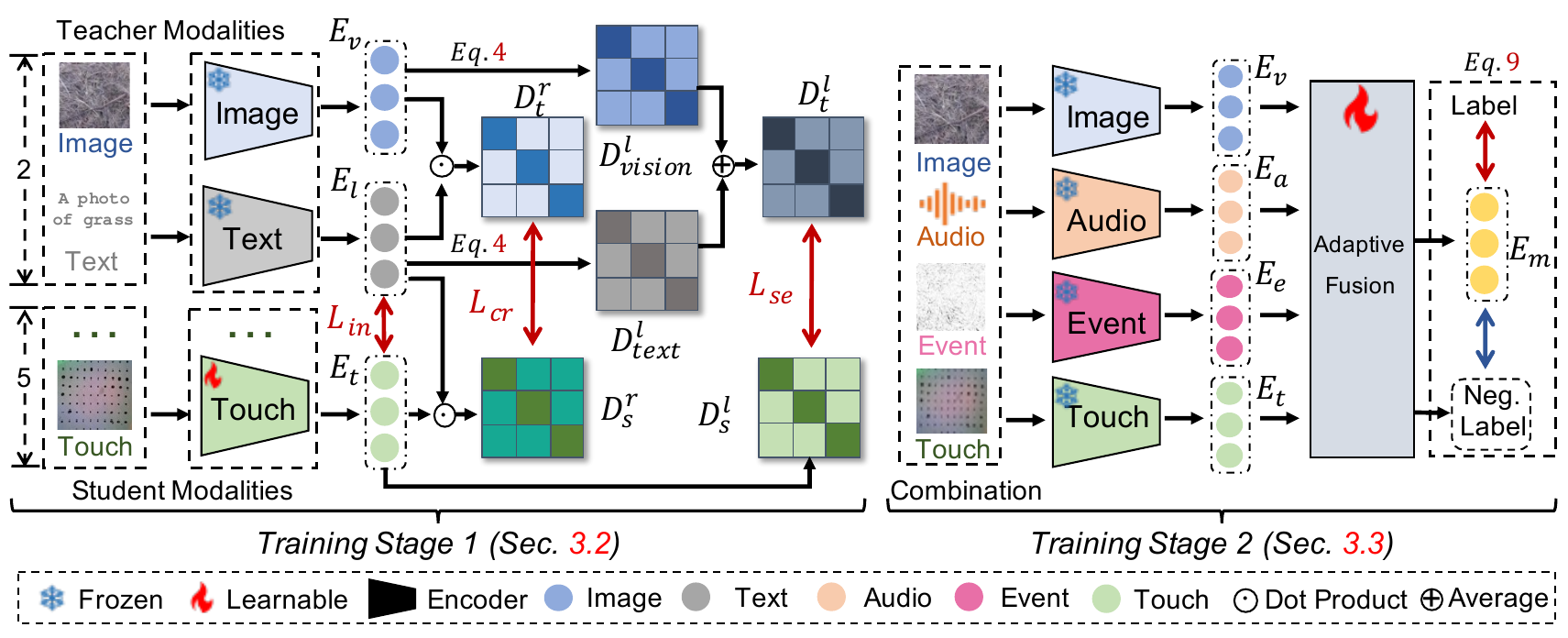}
    \vspace{-10pt}
    \caption{
    \textbf{The overall framework of OmniBind}. We propose a two-stage training approach. \textbf{\textit{Training stage I}}: Aligning the student modalities via CAD module; \textbf{\textit{Training stage II}}: Learning the unified representation space for any modality combination via AF module.
    }
    \label{fig:framework}
\end{figure*}

\section{Related Work}

\noindent \textbf{Multi-modal Foundational Models.}
Existing vision-language models (VLMs)~\cite{fang2021clip2video, mukhoti2023open, zheng2023cvt, cheng2023cico, zhang2021vinvl, radford2021learning,wei2023iclip,li2022blip,li2023blip} demonstrate strong performance in aligning language and image modalities into a shared representation space. Subsequent works extend these models to include additional modalities, \eg, audio~\cite{guzhov2022audioclip,wu2022wav2clip,mahmud2023ave}, point cloud~\cite{zhu2022pointclip,huang2022clip2point,zhang2022pointclip,chen2023clip2scene}, and video~\cite{luo2022clip4clip,clark2017vidloc,lin2022frozen}. 
Inspired by these VLMs, recent works attempt to build a unified representation space for more than three modalities. Generally, these methods can be divided into two categories: 1) multi-encoder-based and 2) universal transformer-based methods. 
For the first line of methods, ImageBind~\cite{han2023imagebind} is a seminal approach that binds all modalities with the image and builds an image-centric representation space. It has been extended to diverse sensing modalities, such as point cloud~\cite{guo2023point} and touch~\cite{yang2024binding}. More recently, LanguageBind~\cite{zhu2023languagebind} and UniBind~\cite{lyu2024unibind} emphasize a language-centric idea and foster the modality-agnostic representation across seven modalities. The universal transformer-based methods~\cite{zhang2024learning,lei2023vit,zhang2023meta} utilize a unified transformer for all modalities via modality-specific tokenizers and prompts.
However, these methods ignore the mismatch and the unequal-scale (data-scale imbalance) problems among modalities between training and inference.
To tackle these problems, our key idea is "modalities help modalities". Intuitively, we propose the CAD to teach the data-constrained modalities to be aligned with the well-trained data-abundant modalities. Then, the AF module is designed to build a unified representation space for any modality combinations.

\noindent \textbf{Modality Fusion.}
Many works attempt to fuse multiple modalities to obtain multi-source information~\cite{su2023recent,duan2022multimodal}. Initial works focus on two-modality fusion, \eg, vision-language~\cite{zhou2020unified,yang2022improving,liu2022universal,bao2021vlmo,bao2022vlmo} or RGB-Depth~\cite{wu2023hidanet,zhao2023arkittrack}, which amalgamate complementary features from both. 
However, these methods often experience significant performance drops when a modality is absent~\cite{zhang2023cmx,liu2024fourier,zhang2022fmcnet,zhang2024learning}. More recent works explore the fusion of more than three modalities~\cite{lyu2024image,tang2024any,wu2023next}, typically using a specific primary modality as a bridge for combining them.
Nonetheless, they struggle to handle any modality combinations flexibly during the inference~\cite{zhang2024learning}. 
To address this, we propose the AF module to fuse the multi-modal embeddings and learn a unified representation space for any modality combinations.

\noindent \textbf{Knowledge Distillation} aims to transfer the knowledge from a teacher model to student model~\cite{hinton2015distilling,wang2020knowledge}. Most KD methods focus within a single modality, distilling logits~\cite{xu2020knowledge,yao2020knowledge}, features~\cite{romero2014fitnets,heo2019comprehensive} and relations~\cite{park2019relational,yang2022cross,zheng2024eventdance} between models. For the cross-modal KD, the correspondence is widely applied to transfer the knowledge~\cite{gupta2016cross,wang2019efficient,zhao2020knowledge,li2020towards}.
In our OmniBind, we introduce the CAD module to align the teacher and student models in modalities with unequal-scale (imbalance) data samples by distilling the intra- and cross-modal correspondence.

\noindent \textbf{Multi-modal Datasets} serve as the foundation for multi-modal learning. Initially, these datasets only consist of visual data and their corresponding categories~\cite{deng2009imagenet,piczak2015esc,jia2021llvip,wu20153d}, which limit their scalability and diversity. To address this issue, the follow-up works pay attention to the abundance of paired dual-modal  datasets~\cite{xu2016msr,chen2015microsoft, kim2019audiocaps, yang2022touch} for the cross-modal retrieval task. Recently, PointBind~\cite{guo2023point} and LanguageBind~\cite{zhu2023languagebind} collect the paired multi-modal datasets, which include more than three modalities. These datasets use language as the bridge to build the \textit{"Text-X"} paired datasets. 
However, these datasets combine multi-modal data by language, which leads to a lack of flexibility in modality combinations. For example, in training on these datasets, the model can't achieve combinations of only visual modalities, \eg, (image, event, and touch).
In light of this, we build the first dataset that consists of seven modalities and enables omni-bind for any of them.

\section{The Proposed OmniBind}
\label{method}

\subsection{Problem Setting and Overview}
\noindent \textbf{Problem Setting:} 
We aim to address the challenges of 1) modality mismatch between training and inference, and 2) data-scale imbalance among modalities. These challenges make it hardly possible for the previous methods, \eg,~\cite{han2023imagebind,lei2023vit} that only utilize a single modality during inference.
As defined in Eq.~\ref{eq:problem}, we consider a function $F(.)$ with multi-modal encoders trained with data samples \(\{M_1, ..., M_N\}\) from a total number of \( N \) modalities, where \(M_1=\{x_{1}^{M_1}, \ldots x_{k}^{M_1}\}\) and $k$ is the batch size. Denote the number of input modalities as \( n \), and it may vary under the following conditions:

\noindent \textbf{1)} When \( n = 1 \), $F(.)$  can provide predictions based on a single modality input during inference;

\noindent \textbf{2)} When \( n > 1 \), $F(.)$  can provide predictions for the inputs with any combination of the \( N \) modalities.
\begin{small}
\begin{equation}
\label{eq:problem}
    f(\{M_1, ... M_N\}) = 
\begin{cases} 
   F(M_i), i \in \{1, \ldots, N\} , & n = 1 \\
   F(Combine(\{M_1, ..., M_n\})), & n > 1
\end{cases}
\end{equation}
\end{small}
where $Combine(.)$ selects any combination of all $N$ modalities $\{M_1, ..., M_n\}$.
This setting requires \( F(\cdot) \) to support adaptive fusion of any combination of input modalities at inference, ranging from 1 to \( n \) modalities. For example, if \( F(\cdot) \) is to encode image (I), touch (T), and event (E) modalities, the possible inference inputs include: I, T, E, I+T, I+E, T+E, and all three combined.

\noindent \textbf{Overview:}
An overview of OmniBind is depicted in Fig.~\ref{fig:framework}.
\textbf{\textit{The key insight}} of our OmniBind is "\textit{Modalities Help Modalities}", which involves teaching data-constrained, \aka, \textit{student}, modalities (audio, point cloud, event, touch and thermal) to be aligned with the well-trained data-abundant, \aka, \textit{teacher}, modalities (image and text). This subtly enables the adaptive fusion of any modalities to build a unified representation space for any combinations. 
Specifically, OmniBind incorporates two training stages: \textbf{I)} We introduce the Cross-modal Alignment Distillation (CAD) module (Sec. \ref{sec:3.1}), which aligns the student modalities with the teacher modalities;
\textbf{II)} We propose the Adaptive Fusion (AF) module (Sec. \ref{sec:3.2}) to fuse modalities, thereby building a unified representation space for any modality combinations. Moreover, we compile and align data from existing datasets of seven modalities with the support of the LLM~\cite{openai2023gpt4} and MLLM~\cite{zhang2023llama} to construct the modality-free dataset encompassing any modality combinations (Sec. \ref{sec:3.3}) for training and inference.


\subsection{Cross-modal Alignment Distillation (CAD)}
\label{sec:3.1}
In Stage I, our CAD module aims to address the data-scale imbalance problem between student and teacher modalities and effectively align the student modalities with teacher modalities through knowledge distillation (KD), as shown in Fig.~\ref{fig:framework}.
Specifically, we randomly sample one, \eg, touch, from the five student modalities. The sampled modality is then aligned with the teacher modalities, \ie, image and text. After this, another modality, \eg, event, is randomly sampled from the remaining four student modalities, and aligned with the teacher modalities. This cross-modal distillation process is iterated for five times in total. The sampled student modality is aligned with the teacher modalities using the same optimization strategy. Take touch as an example,  as shown in Fig.~\ref{fig:framework}, its embeddings $E_t=\{E_{t_i}, ..., E_{t_k}\}$ are extracted using a learnable touch encoder. 

For the teacher modalities, initially, we freeze the image and text encoders which are pre-trained with large-scale data samples as teacher models.
The image and text embeddings are extracted from the image \( \{x_{1}^{img}, \ldots, x_{k}^{img} \} \) and text \( \{x_{1}^{txt}, \ldots, x_{k}^{txt} \} \) input, $E_v =\{E_{v_i}, ..., E_{v_k}\}$ and $E_l=\{E_{l_i}, ..., E_{l_k}\}$ for distillation, respectively.  Following~\cite{girdhar2023imagebind, lyu2024unibind}, we employ the extracted text embeddings as the supervisory signal for KD with contrastive learning:
\begin{small}
\begin{equation}
    \mathcal{L}_{in} = -\log\frac{\exp(E_{t_i} \cdot E_{l_i}^{T}/\tau)}{\exp(E_{t_i} \cdot E_{l_i}^{T}/\tau) + \sum_{j \ne i}(\exp(E_{t_i} \cdot E_{l_j}^{T}/\tau))},
\end{equation}
\end{small}
To effectively train the touch modality, we distill knowledge in two aspects: 1) the cross-correspondence between embeddings from the teacher and student modalities, and 2) the self-correspondence of the extracted embeddings from each modality. 
For the former, we first calculate the cross-correspondence $D_{t}^{r}$ of the teacher modalities, \ie, using text and image embeddings $E_v =\{E_{v_i}, ..., E_{v_k}\}$ and $E_l=\{E_{l_i}, ..., E_{l_k}\}$:
\begin{equation}
    D_{t}^{r} = \{E_{l_i}, ..., E_{l_k}\} \cdot {\{E_{t_i}, ..., E_{t_k}\}}^{T},
\end{equation}
Then, we transfer the cross-modality knowledge $D_{t}^{r}$ to touch modality using the \textit{Kullback-Leibler} (KL) divergence loss:
\begin{small}
\begin{equation}
    \mathcal{L}_{cr} = \sum_{} D_{t}^{r} \log\left(\frac{D_{t}^{r}}{\{E_{l_i}, ..., E_{l_k}\} \cdot {\{E_{t_i}, ..., E_{t_k}\}}^{T}}\right),
\end{equation}
\end{small}

Afterwards, we calculate the self-correspondence of the image modality $D_{img}^{l}$ and text modality $D_{txt}^{l}$ via inner-product, and average $D_{img}^{l}$ and $D_{txt}^{l}$ to get $D_{t}^{l}$:
\begin{equation}
    D_{t}^{l} = (D_{img}^{l} \cdot {D_{img}^{l}}^{T} + D_{txt}^{l} \cdot {D_{txt}^{l}}^{T}) / 2,
\end{equation}
We then regard the averaged self-correspondence $D_{t}^{l}$ as the intra-modality knowledge for the touch modality.
The KL divergence is applied as the loss for the intra-modality knowledge transfer:
\begin{small}
\begin{equation}
    \mathcal{L}_{se} = \sum_{} D_{t}^{l} \log\left(\frac{D_{t}^{l}}{\{E_{t_i}, ..., E_{t_k}\} \cdot {\{E_{t_i}, ..., E_{t_k}\}}^{T}}\right).
\end{equation}
\end{small}

\begin{figure}[t!]
    \centering
    \includegraphics[width=0.9\linewidth]{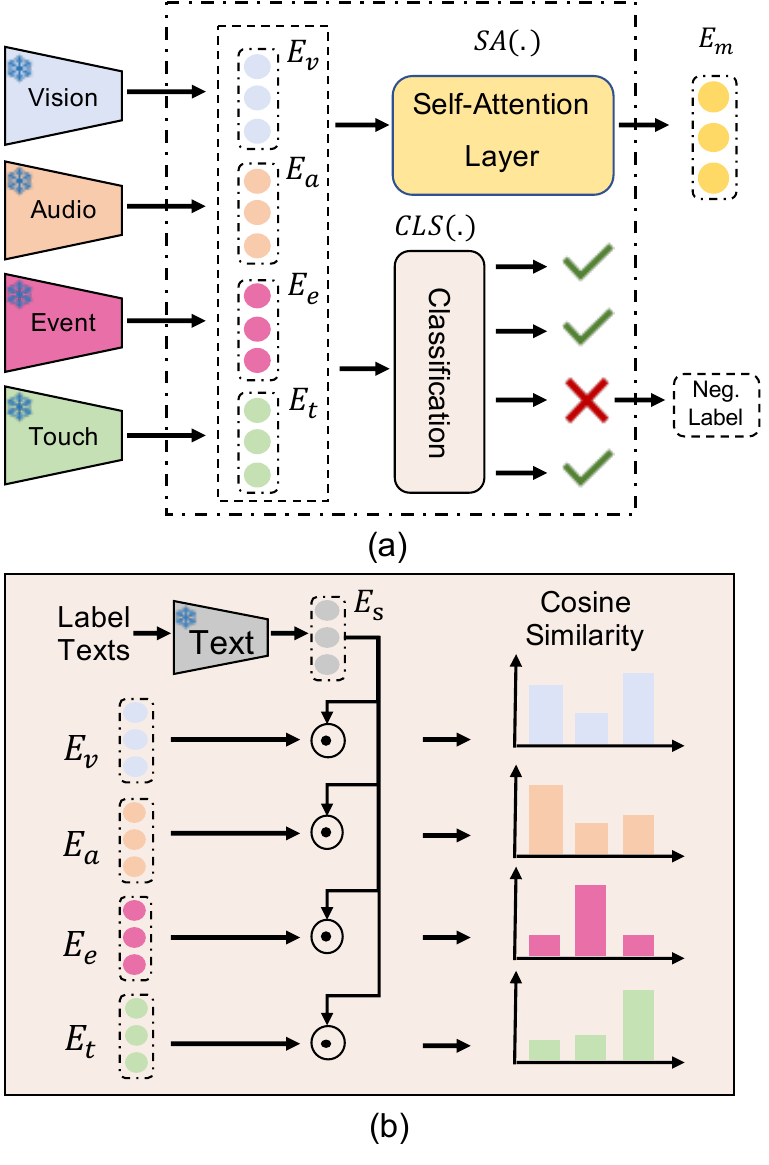}
    \vspace{-6pt}
    \caption{
    The Adaptive Fusion module. \textbf{(a)} The framework of our proposed AF module; \textbf{(b)} The details of the classification operation in the AF module.
    }
    \vspace{-6pt}
    \label{fig:adpater}
\end{figure}

The overall loss consists of:
\begin{equation}
    \mathcal{L}_{stage 1} = \mathcal{L}_{in} + \mathcal{L}_{cr} + \mathcal{L}_{se}.
\end{equation}
This is the CAD module for the randomly sampled modality, \ie, touch, and the same CAD module is performed for the other four student modalities.
The proposed CAD module effectively reinforces student modalities, \ie, touch, to be aligned with the teacher modalities, \ie, text, by cross-modality and intra-modality knowledge. This alleviates the data-scale imbalance problems and aligns all the student modalities to the teacher modalities, paving the way to learning a unified representation space for training stage II.



\subsection{Adaptive Fusion (AF)}
\label{sec:3.2}
After achieving the alignment between all student 
and teacher modalities, the AF module in Stage II, is designed to build a unified representation space for any modality combinations. The detailed structure of the AF module is depicted in Fig.~\ref{fig:adpater}.
Here, the parameters of all modalities' encoders are frozen, including the teacher and student modalities. 
Specifically, we first randomly sample any combinations from all the modalities as inputs for the AF module. For example, we sample image, audio, event, and touch as input for the AF module, as shown in Fig.~\ref{fig:adpater} (b).

Then the label embeddings \( E_{s}=\{ E_{s_1}, ..., E_{s_k}\}\) are extracted by passing all the label texts into the text encoder. 
The multi-modal embeddings, \ie, $E_v, E_a, E_e, E_t$ in Fig.~\ref{fig:adpater} (a), are obtained by passing any combination inputs to the corresponding multi-modal encoders.
Afterwards, the multi-modal embeddings are passed to the ‘Classification’ operation $CLS(.)$ in Fig.~\ref{fig:adpater} (b), to find the output predictions for each modality.
Concretely, we calculate the cosine similarity between all the label embeddings \( E_{s}=\{ E_{s_1}, ..., E_{s_k}\}\) and the multi-modal embeddings $E_v, E_a, E_e, E_t$. For instance, the final prediction of input from vision modality is the corresponding label with the highest similarity score in \(Sim(E_v, E_s)\). By doing so, we get the outputs for each modality in the combinations.



Subsequently, we use the real label as the benchmark to find the correct and incorrect predictions of all modalities. The incorrect predictions are used as negative supervision for improving the overall robustness of OmniBind. 
Actually, there is always at least one incorrect prediction among multi-modal inputs in experiments, and in case there is no incorrect prediction, AF module randomly picks one as a negative label for contrastive learning.
Specifically, 
we fuse the multi-modal inputs using a self-attention layer $SA(.)$ to obtain the final multi-modal embedding $E_m$ for predictions:
\begin{equation}
    E_m = \text{mean}(SA({E_v, ..., E_t})),
    \end{equation}
    where $\text{mean}(.)$ is the averaging function.

The contrastive loss for improving the overall robustness in Stage II can be expressed as:
\begin{small}
\begin{equation}
    \mathcal{L}_{stage 2} = -\log\frac{\exp(E_{m} \cdot E_{pos}^{T}/\tau)}{\exp(E_{m} \cdot E_{pos}^{T}/\tau) + \sum_{j \ne i}(\exp(E_{m} \cdot E_{neg}^{T}/\tau))},
\end{equation}
\end{small}
where $E_{pos}$ is the embeddings of {\fontfamily{qcr}\selectfont [a photo of a "label"]} and $E_{neg}$ is the embeddings of {\fontfamily{qcr}\selectfont [a photo of a "negative label"]} extracted by the text encoder, as shown in Fig.~\ref{fig:framework}.
Overall, the AF module handles any modality combination inputs flexibly and exhibits robust performance across any modality combinations. 



\begin{figure}[t!]
  \centering
  \includegraphics[width=\linewidth]{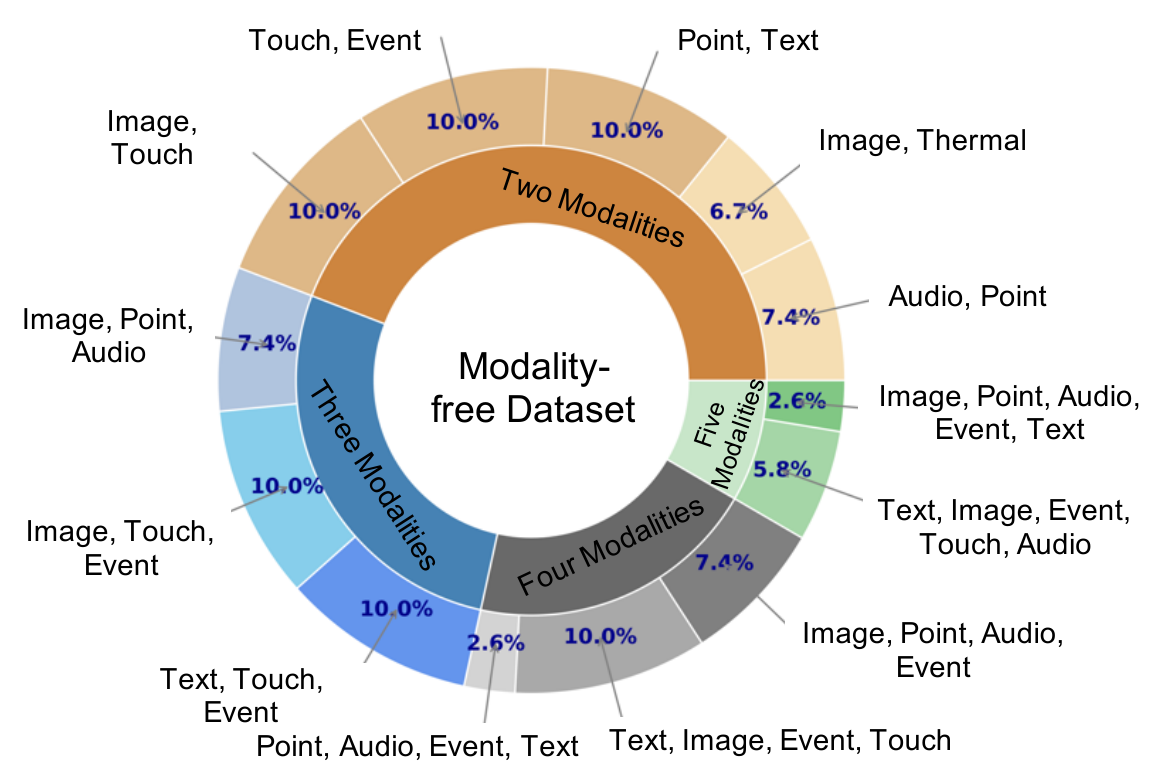}
    \vspace{-6pt}
    \caption{
    Overview of the modality-free dataset.
    }
    \vspace{-6pt}
    \label{fig:dataset_details}
\end{figure}

\begin{table}[t!]
\renewcommand{\tabcolsep}{3pt}
\resizebox{\linewidth}{!}{
\begin{tabular}{l|cccc} 
\toprule
\multicolumn{1}{c|}{Dataset Details} & Modality &  Scale& \#cls & Metric\\ \midrule
ImageNet-1K (IN-1K)~\cite{deng2009imagenet} & Image & 1,280K & 1,000 & Acc \\
Caltech-101 (cal)~\cite{fei2004learning} & Image & 8K & 101 & Acc \& Recall \\
ShapeNet-part (ShapeNet)~\cite{wu20153d} &Point Cloud & 16K & 16 & Acc  \\
ESC 5-folds (ESC)~\cite{piczak2015esc}& Audio & 2K & 50 & Acc  \\
Urban-Sound-8K (Urban-S)~\cite{salamon2014dataset}& Audio & 8K & 10 & Acc  \\
LLVIP (LLVIP) ~\cite{jia2021llvip} & Thermal & 15K & 2 & Acc \\
N-ImageNet-1K (N-IN-1K)~\cite{kim2021n} & Event & 1,280K & 1,000 & Acc  \\
N-Caltech-101 (N-cal)~\cite{orchard2015converting} & Event & 8K & 101 & Acc \& Recall  \\
Touch-and-Go (TouchGo)~\cite{yang2022touch} & Touch & 13.9K & 20 & Acc  \\
\bottomrule
\end{tabular}}
\vspace{-6pt}
\caption{The details of source datasets in our modality-free dataset.}
\vspace{-6pt}
\label{tab: datasets}
\end{table}

\begin{table*}[t!]
\renewcommand{\tabcolsep}{12pt}
\resizebox{\linewidth}{!}{
\begin{tabular}{l|ccccc} 
\toprule
\multicolumn{1}{l|}{Method}&  \multicolumn{5}{c}{Combinations of Two Modalities}\\ \cmidrule{2-6}
\multicolumn{1}{c|}{} &  Point+Text&  Image+Touch&  Image+Thermal&  Event+Touch&  \multicolumn{1}{c}{Point+Audio}\\ \midrule
ImageBind~\cite{girdhar2023imagebind}+Mean & \tiny{\XSolidBrush} & \tiny{\XSolidBrush} & 65.97 & \tiny{\XSolidBrush} & \tiny{\XSolidBrush} \\ 
PointBind~\cite{guo2023point}+Mean & 86.20 & \tiny{\XSolidBrush} & 65.97 & \tiny{\XSolidBrush} & 61.66 \\ 
LanguageBind~\cite{zhu2023languagebind}+Mean & \tiny{\XSolidBrush} & \tiny{\XSolidBrush} & 89.70 & \tiny{\XSolidBrush} & \tiny{\XSolidBrush}  \\ 
UniBind~\cite{lyu2024unibind}+Mean & 87.10 & 42.60 & 71.02 & 17.00 & 63.19  \\ 
UniBind~\cite{lyu2024unibind}+Unseen~\cite{zhang2024learning} & 82.50 & 82.12 & 85.72 & 79.40 & 90.10  \\ 
UniBind~\cite{lyu2024unibind}+Linear& 87.10 & 75.60 & 87.60 & 68.60 & 96.24  \\ 
Ours & \textbf{95.00} & \textbf{89.40} & \textbf{97.90} & \textbf{85.60} & \textbf{97.85} \\ 
\rowcolor{gray!10}$\Delta $ &\textbf{+7.90} &\textbf{+7.28} &\textbf{+10.30} & \textbf{+6.20} & \textbf{+1.61}  \\
\bottomrule
\end{tabular}}
\vspace{-6pt}
\caption{Results with the two-modality combinations. \XSolidBrush: not support.} 
\label{tab: compare_2m}
\end{table*}

\begin{table*}[t!]
\renewcommand{\tabcolsep}{14pt}
\resizebox{\linewidth}{!}{
\begin{tabular}{l|cccccc|c} 
\toprule
Method&  \multicolumn{2}{|c}{Three Modalities} & \multicolumn{2}{c}{Four Modalities} & \multicolumn{2}{c|}{Five Modalities} & \\ \cmidrule{2-7}
&  \multicolumn{1}{|c}{C1}& C2& C3& C4& C5& \multicolumn{1}{c|}{C6} & Total \\ \midrule
PointBind~\cite{guo2023point}+Mean & 68.63 & \tiny{\XSolidBrush} & \tiny{\XSolidBrush} & \tiny{\XSolidBrush} & \tiny{\XSolidBrush} & \tiny{\XSolidBrush} & -\\ 
UniBind~\cite{lyu2024unibind}+Mean & 69.33 & 33.60 & 25.80 & 69.16 & 37.58 & 90.90 & 48.37\\ 
UniBind~\cite{lyu2024unibind}+Unseen~\cite{zhang2024learning} & 88.70 &  82.40 & 78.05 & 92.85 & 81.20 & 97.25 & 85.01 \\
UniBind~\cite{lyu2024unibind}+Linear & 94.10 & 80.80 & 81.40 & 97.31 & 90.00 & 98.24 & 87.84 \\ 
Ours & \textbf{96.78} &\textbf{ 90.40} & \textbf{89.60} & \textbf{98.65} & \textbf{92.75} & \textbf{98.48} & \textbf{92.42} \\ 
\rowcolor{gray!10}$\Delta $ &\textbf{+2.68} &\textbf{+9.60} &\textbf{+8.20} &\textbf{+1.34} &\textbf{+2.75} &\textbf{+0.24 }&\textbf{+4.05} \\
\bottomrule
\end{tabular}}
\vspace{-6pt}
\caption{Results with the combinations (C) of three, four, and five modalities. Image: I, Audio: A, Point Cloud: P, Text: Te, Event: E, and Touch: T. C1: \{I, A, P\}; C2: \{Te, E, T\}; C3: \{Te, E, T, I\}; C4: \{I, E, A, P\}; C5: \{Te, E, T, I, A\}; C6: \{I, P, A, E, Te\}.}
\label{tab: compare_345m}
\end{table*}

\begin{table*}[t!]
\renewcommand{\tabcolsep}{16pt}
\begin{minipage}[b]{0.30\linewidth}
\centering
\resizebox{\linewidth}{!}{
\begin{tabular}{lc}
\toprule
\multicolumn{1}{l}{Model} & \multicolumn{1}{c}{ACC} \\ \midrule
VT CMC~\cite{yang2022touch} & 47.10 \\
SSVTP~\cite{kerr2022self} & 47.60 \\
VT CMC-all~\cite{yang2022touch} & 49.20 \\
SSVTP-all~\cite{kerr2022self} & 43.80 \\
UniTouch~\cite{yang2024binding} & 61.30 \\
ViT-Lens~\cite{lei2023vit} & 63.11 \\
Ours & \textbf{67.45} \\
\rowcolor{gray!10}$\Delta $ & \textbf{+4.34} \\
\bottomrule
\end{tabular}}
\vspace{-4pt}
\caption{Results on touch.}
\label{tab: compare_touch}
\end{minipage}\hfill
\renewcommand{\tabcolsep}{16pt}
\begin{minipage}[b]{0.30\linewidth}
\centering
\resizebox{\linewidth}{!}{
\begin{tabular}{lc}
\toprule
\multicolumn{1}{l}{Model} & \multicolumn{1}{c}{ACC} \\ \midrule
RG-CNN~\cite{bi2019graph} & 65.70 \\
EST~\cite{gehrig2019end} & 81.70 \\
DVS-ViT~\cite{wang2022exploiting} & 83.00 \\
UniBind~\cite{lyu2024unibind} & 78.05 \\
EventCLIP~\cite{wu2023eventclip} & 93.57 \\
EventBind~\cite{zhou2023clip} & 93.74 \\
Ours & \textbf{94.60} \\
\rowcolor{gray!10}$\Delta $ & \textbf{+0.86} \\
\bottomrule
\end{tabular}}
\vspace{-4pt}
\caption{Results on event.}
\label{tab: compare_event}
\end{minipage}\hfill
\renewcommand{\tabcolsep}{16pt}
\begin{minipage}[b]{0.34\linewidth}
\centering
\resizebox{\linewidth}{!}{
\begin{tabular}{lc}
\toprule
\multicolumn{1}{l}{Model} & \multicolumn{1}{c}{ACC} \\ \midrule
ImageBind~\cite{girdhar2023imagebind} & 63.40 \\
UniBind~\cite{lyu2024unibind} & 64.67 \\
OpenCLIP~\cite{cherti2023reproducible} & 82.20 \\
LanguageBind~\cite{zhu2023languagebind} & 88.91 \\
Ours & \textbf{92.60} \\
\rowcolor{gray!10}$\Delta $ & \textbf{+3.69} \\
\bottomrule
\end{tabular}}
\vspace{-4pt}
\caption{Results on thermal.}
\label{tab: compare_thermal}
\end{minipage}
\vspace{-6pt}
\end{table*}

\subsection{Modality-free Dataset}
\label{sec:3.3}
The goal of the proposed modality-free dataset is to address the mismatch problem between training and inference and also evaluate the effectiveness of our OmniBind.
We construct the dataset consisting of any modality combinations for training and inference. We initially align the labels of multiple datasets, as enumerated in Tab.~\ref{tab: datasets}, powered by LLM~\cite{openai2023gpt4}. 
Then we generate the descriptions for all the data samples via MLLM~\cite{zhang2023llama} and match the data samples with the same semantics of the descriptions.
As depicted in Fig.~\ref{fig:dataset_details}, our dataset consists of \textit{50,080} samples of modality combinations, with the most prevalent combinations being two modalities, accounting for 46\%. The proportions for three-modal, four-modal, and five-modal combinations are 15\%, 16\%, and 17\%, respectively. \textit{For more details of our dataset, please refer to the appendix. }
\section{Experiments and Evaluation}
\label{experiment}

\subsection{Experimental Setup}


\noindent \textbf{Datasets:}
We compile our modality-free dataset from nine datasets. The details of these datasets are presented in Tab.~\ref{tab: datasets}. We show more details \textit{in the appendix}. Additionally, we validate our OmniBind on three single-modality datasets and follow the benchmarks established by UniBind~\cite{lyu2024unibind} and UniTouch~\cite{yang2024binding}. Our performance on the student modalities is compared with existing baselines and state-of-the-art methods in Tab.~\ref{tab: compare_touch}, \ref{tab: compare_event}, and \ref{tab: compare_thermal}.

\noindent \textbf{Backbone Models:}
We employ visual encoders from UniBind~\cite{lyu2024unibind} for the image, point cloud, and audio modalities. For the event and thermal modalities, we enhance the existing UniBind encoders by adding an MLP (multi-layer perception) layer at the end. During training, we configure only the parameters of these MLP layers as learnable. Similarly, for the touch modality, we utilize UniBind’s image encoder augmented with an MLP layer at its end, setting only the MLP layer's parameters as learnable.

\noindent \textbf{Implementation Details:}
In the training process, our OmniBind is implemented in two stages. During CAD in Stage I, we freeze the parameters of the teacher modality encoders and set only the parameters of the student modality encoders as learnable. This strategy effectively facilitates the transfer of knowledge from the teacher to the student modalities. In Stage II, we freeze the parameters of all the multi-modal encoders and only train the adaptive fusion module to achieve a unified representation space for any modality combinations. For comparison experiments, we employ baseline methods such as ImageBind~\cite{han2023imagebind}, PointBind~\cite{guo2023point}, LanguageBind~\cite{zhu2023languagebind}, and UniBind~\cite{lyu2024unibind}. As they cannot be used to accept any modality combination inputs, we add an \textbf{average fusion layer at the end} of them for fusing the modality embeddings. Additionally, we introduce two baseline models: UniBind+Linear, which incorporates a linear layer at the end of UniBind for recognizing multi-modal combinations; and UniBind+Unseen, which adds an unseen module from~\cite{zhang2024learning} to the end of UniBind.
Furthermore, we compare our OmniBind with state-of-the-art methods for single modalities, such as UniTouch~\cite{yang2024binding} and EventBind~\cite{zhou2023clip}.

\subsection{Results under Arbitrary Modality Combination Setting}
We divide the comparison experiments into two categories according to the number of total modalities: two-modality combinations and more than three-modality combinations.

\noindent \textbf{Two-modality Combinations.}
In Tab.~\ref{tab: compare_2m}, we present the experimental results for combinations of two modalities with our modality-free dataset. OmniBind demonstrates \textit{\textbf{robust performance}} in recognizing combinations consisting of two modalities. Notably, OmniBind achieves \textit{\textbf{an average gain of 6.52\% across all two-modality combinations}}, including pairs such as (point, text), (image, touch), (image, thermal), (event, touch), and (point, audio). Specifically, OmniBind exhibits exceptional performance with combinations of student modalities, achieving a \textbf{6.20\%} gain for the combinations of event and touch modalities. These robust results demonstrate the effectiveness of the CAD module in transferring knowledge from teacher modalities and crossing the data-scale gap between teacher and student modalities.

\noindent \textbf{More than Three-modality Combinations.}
In Tab.~\ref{tab: compare_345m}, we illustrate OmniBind’s capability in processing combinations comprising more than three modalities. OmniBind achieves \textit{\textbf{an average improvement of +4.05\%}}. Particularly, OmniBind records more substantial gains with combinations involving more student modalities, such as a \textbf{9.60\%} improvement with the combination of text, event, and touch modalities. This performance highlights the strengths of our approach in aligning student modalities and demonstrates its flexibility and effectiveness in recognizing and balancing combinations of more than three modalities.


\vspace{-5pt}
\subsection{Results under Single-modality Setting}

\noindent \textbf{Touch:}
As shown in Tab.~\ref{tab: compare_touch}, Compared to the state-of-the-art (SoTA) methods, OmniBind secures a significant improvement of +4.34\% in the material recognition benchmark~\cite{yang2022touch}. This gain highlights the superiority of OmniBind in single-modality recognition tasks.\\
\noindent \textbf{Event:}
The quantitative results of OmniBind in the event modality is outlined in Tab.~\ref{tab: compare_touch}. OmniBind achieves SoTA performance on the N-Caltech dataset~\cite{orchard2015converting} while reducing \textbf{90\%} of the learnable parameters compared with EventBind~\cite{zhou2023clip}, the latest SoTA method in the event-based vision. \\
\noindent \textbf{Thermal:}
For the thermal modality, we compare the performance of OmniBind against experimental baselines. 
The recognition results on the LLVIP dataset~\cite{jia2021llvip} are reported, adhering to the benchmarks set by ImageBind~\cite{han2023imagebind}. OmniBind demonstrates strong performance, achieving a 3.69\% gain on the thermal recognition benchmark~\cite{han2023imagebind}.

\section{Ablation Study}

\begin{table}[t!]
\centering
\renewcommand{\tabcolsep}{8pt}
\resizebox{\linewidth}{!}{
\begin{tabular}{l|c|ccc}
\toprule
\multicolumn{1}{l|}{Loss} & \multicolumn{1}{c|}{Align Center} & \multicolumn{1}{c}{Touch} & \multicolumn{1}{c}{Event} & \multicolumn{1}{c}{Thermal} \\ \midrule
$L_{in}$ & & 61.79 & 85.10& 87.10  \\
$L_{in}$ + $L_{cr}$ & Text & 62.70 & 91.66 & 92.10 \\
$L_{in}$ + $L_{se}$ & & 62.31 & 88.20 & 91.50 \\
All & & \textbf{67.45} & \textbf{94.60} & \textbf{92.60} \\ \midrule
$L_{in}$ & & 48.68 & 78.90 & 58.72 \\
$L_{in}$ + $L_{cr}$ & Image & 55.20 & 84.10 & 61.07 \\
$L_{in}$ + $L_{se}$ & & 53.75 & 83.20 &59.92 \\
All & & \textbf{59.83} & \textbf{86.90}& \textbf{63.10}\\
\bottomrule
\end{tabular}}
\vspace{-2pt}
\caption{Ablation of CAD's losses.}
\label{tab: ab_loss}
\vspace{-6pt}
\end{table}

\begin{figure}[t!]
    \centering
    \includegraphics[width=\linewidth]{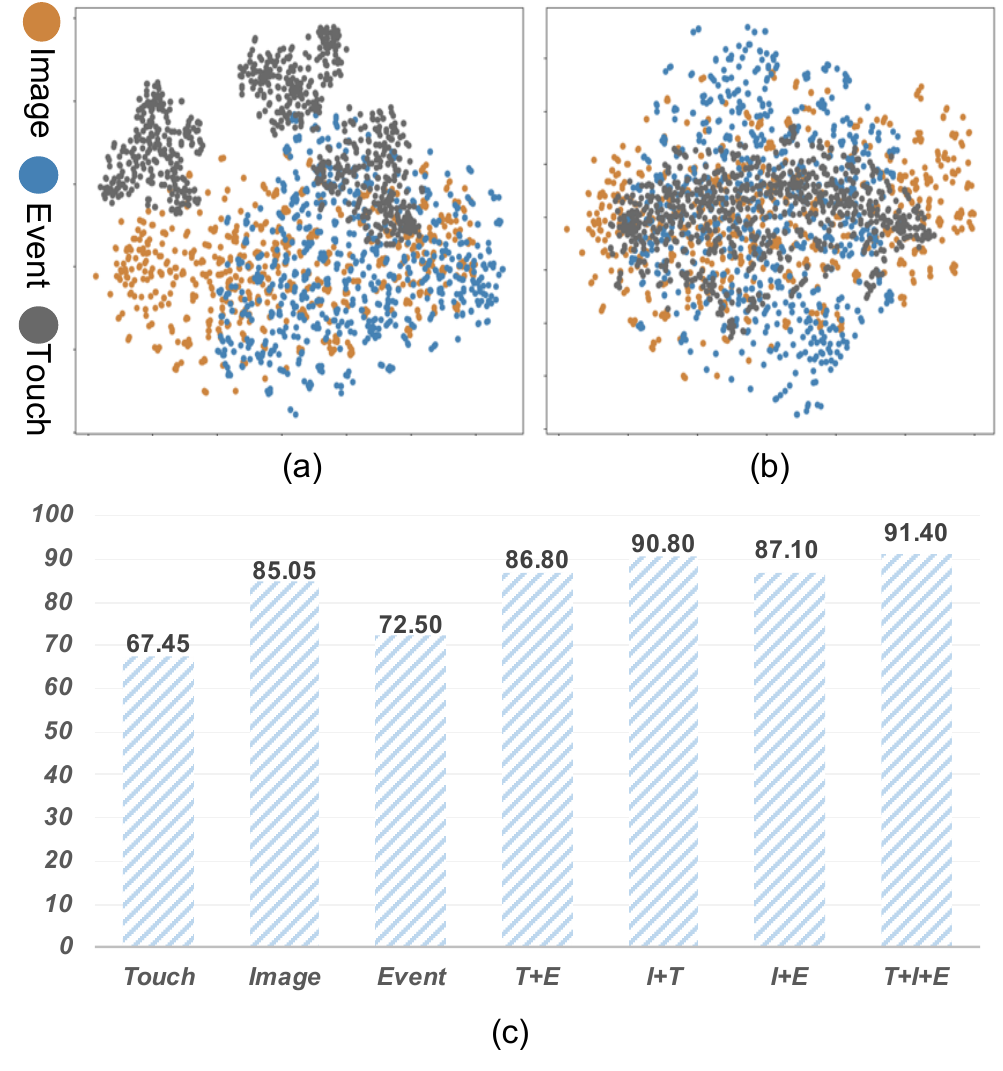}
    \vspace{-10pt}
    \caption{
    The t-SNE visualization (a) without CAD and (b) with CAD. (c): the ablation study of the modality numbers.
    }
    \vspace{-6pt}
    \label{fig:ab}
\end{figure}

\noindent \textbf{Effectiveness of the proposed loss functions.} To evaluate the effectiveness of our proposed loss functions: $L_{in}$, $L_{cr}$, and $L_{se}$ in CAD, we conduct ablation studies in the touch, event, and thermal modalities with image and text as align centers, respectively. The results, outlined in Tab.~\ref{tab: ab_loss}, demonstrate that all of these loss functions significantly improve the overall performance in three modalities. 
The cross-modal knowledge transferred by $L_{cr}$ and the intra-modal knowledge transferred
by $L_{se}$ can be effective in improving performance and helping student modalities align to the teacher modalities' representation space. 
Overall, the integration of all three proposed loss functions culminate in the highest performance, \ie, \textit{\textbf{67.45\% Acc for the touch modality, 94.60\% Acc for the event modality, and 92.60\% Acc for the thermal modality.}} These results show the superiority of all loss functions in aligning student modalities with teacher modalities. We also show the \textit{\textbf{t-SNE visualization}} of the representations of image, touch, and event modalities in Fig.~\ref{fig:ab} (a) and (b). Obviously, using our CAD makes the embeddings from different modalities become more clustered. The visualization results show that the proposed CAD can effectively align embeddings from different modalities in the high-level representation space.


\noindent \textbf{Ablation of AF Module.}
As shown in Tab.~\ref{tab: ab_stage2}, we show the efficiency of the $SA(\cdot)$ and the $CLS(\cdot)$ in the AF module. We conduct the experiments with any modality combinations, ranging from 2 to 5 modalities. Note that the results in Tab.~\ref{tab: ab_stage2} are the average of all possible combinations. Both $CLS(\cdot)$ and $SA(\cdot)$ play positive roles in improving the overall performance, and using them simultaneously achieves the highest overall accuracy of 92.42\%.
Besides, we also assess the effectiveness of the $SA(\cdot)$ layer in Tab.~\ref{tab: atten}. Evidently, replacing our $SA(\cdot)$ with an outer product or linear layer leads to significant performance drops. This indicates the superiority of the $SA(\cdot)$ in AF module.

\begin{table}[t!]
    \centering
    \renewcommand{\tabcolsep}{6pt}
    \resizebox{\linewidth}{!}{
    \begin{tabular}{l|cccc|c} 
    \toprule
    \multicolumn{1}{l|}{Method} & 2M & 3M & 4M & 5M & Total\\ \midrule
    $Cat$ & 56.18 & 42.88 & 61.95 & 64.24 & 48.37  \\ 
    $CLS(\cdot)$ & 92.54 & \textbf{91.03} & 92.29 & 93.45 & 91.07  \\ 
    \rowcolor{gray!10} $CLS(\cdot)+SA(\cdot)$ & \textbf{93.93} & 90.68 & \textbf{94.98} & \textbf{95.10} & \textbf{92.42} \\
    \bottomrule
    \end{tabular}}
    \vspace{-6pt}
    \caption{Ablation study of the AF module. $Cat$ means directly concatenating all the multi-modal embeddings for final predictions. $SA(\cdot)$ stands for the self-attention layer. (Modality: M)}
    \label{tab: ab_stage2}
\end{table}

\noindent \textbf{Robustness of Adaptive Fusion Module.}
To assess the robustness of our AF module, we adopted the modality-noise setting proposed by Zhang et al.\cite{zhang2024learning} to evaluate our approach. We randomly selected one input modality and introduced noise to it. As reported in Tab.\textcolor{red}{11} in the appendix, OmniBind training with negative labels led to \textit{\textbf{a notable average improvement of +3.74\%}}. This enhancement underscores the crucial role of training with negative labels in boosting the robustness of OmniBind for modality combination recognition. Training with negative labels shows stronger performance gains with combinations of three modalities compared to those of four and five modalities, illustrating the advantages of our method in resisting a broader range of noise interference.

\begin{table}[t!]
    \centering
    \renewcommand{\tabcolsep}{6pt}
    \resizebox{\linewidth}{!}{
    \begin{tabular}{l|cccc|c} 
    \toprule
    \multicolumn{1}{l|}{Method} & 2M & 3M & 4M & 5M & Total\\ \midrule
    Outer Product & 22.50 & 28.91 & 41.20 & 44.82 & 33.67 \\ 
    Linear & 83.03 & 87.36 & 92.65 & 94.14 & 87.84   \\ 
    \rowcolor{gray!10} $SA(\cdot)$ (Ours) & \textbf{93.93} & 90.68 & \textbf{94.98} & \textbf{95.10} & \textbf{92.42} \\
    \bottomrule
    \end{tabular}}
    \vspace{-6pt}
    \caption{Ablation study on the self-attention of AF module. Outer product and linear means we replace the $SA(\cdot)$ with outer product operation and a linear layer. }
    \label{tab: atten}
    \vspace{-6pt}
\end{table}

\section{Discussion}
\noindent \textbf{Modality Combinations:} We investigate the effectiveness of including more modalities in the combination on our dataset. As shown in Fig.~\ref{fig:ab} (c). OmniBind trained with three modalities exhibited the best performance, and training with two modalities showed significant improvement over training with a single modality. These results reveal the significance of our key idea "Modality Help Modality" and also show the importance of our Modality-free dataset.  

\noindent \textbf{Modality-free Dataset:} In constructing the dataset, we employ coarse alignment at the label level and fine-grained alignment at the description level for assembling and collecting the multi-modal combined data. The data obtained from this method are not perfectly aligned, but they do ensure semantic alignment. For recognition tasks, this approach is justifiable and aligns well with practical application scenarios where perfect alignment of multi-modal sensors cannot be guaranteed.


\noindent \textbf{Broader Impacts:} OmniBind enhances autonomous system adaptability, enabling robust performance in dynamic environments, which is beneficial for industrial automation and accessibility. However, its reliance on diverse modalities may increase complexity and resource demands. 

\section{Conclusion}

In this paper, we proposed OmniBind, a universal multi-modal learning approach that crosses imbalance among multiple modalities and learns unified representations for any combination of modalities. 
\textbf{\textit{The key insight}} of our OmniBind is "\textit{Modalities Help Modalities}," which involves teaching data-constrained, \ie, student, modalities (\eg, touch and thermal) in alignment learning through well-trained data-abundant, \ie, teacher, modalities (\eg, image and texts) and further enables the flexible learning of unified representations for any combination of modalities. 

\noindent \textbf{Limitations and Future Works.}
There are still downstream applications of the free combinations of multiple modalities to be explored. In response, our future work will concentrate on applying OmniBind on more downstream multi-modal tasks.



\clearpage
{
    \small
    \bibliographystyle{plain}
    \bibliography{main}
}


\clearpage


\appendix

\section*{Appendix}
\label{appendix}

\section{More Details of Datasets}
\subsection{More Details of Source Datasets}

\noindent \textbf{ImageNet-1K (IN-1K)}~\cite{deng2009imagenet}.
This dataset is a fundamental image repository designed for recognition tasks, encompassing over 1,000 categories. It is utilized for both training and evaluation purposes. In the zero-shot setting, we assess both baseline models and our proposed method on the test set without prior training, employing accuracy as the metric for evaluation.

\noindent \textbf{Caltech-101 (Cal)} ~\cite{fei2004learning}.
The Caltech-101 dataset is an established benchmark in computer vision, specifically designed for object recognition tasks. It comprises images from 101 diverse object categories, reflecting a variety of real-world scenes. In this study, we use the dataset for model evaluation in object recognition and extend its use to cross-modal retrieval tasks involving Caltech-101 and N-Caltech-101. Accuracy is the metric for recognition.

\noindent \textbf{ShapeNet-part (ShapeNet)}~\cite{wu20153d}.
ShapeNet-part is a key benchmark dataset widely used for 3D segmentation tasks. In this work, we frame the evaluation on ShapeNet-part as a recognition task. The dataset includes 16 categories of 3D objects, with accuracy as the chosen evaluation metric.

\noindent \textbf{ESC 5-folds (ESC)}~\cite{piczak2015esc}.
This dataset includes 2,000 audio clips, each lasting five seconds and classified into 50 distinct categories. In the zero-shot setting, we utilize the entire dataset to assess both baseline models and our proposed method. For fine-tuning, models are trained on the designated training set and evaluated on the test set, using accuracy as the evaluation metric.

\noindent \textbf{Urban-Sound-8K (Urban-S)}~\cite{salamon2014dataset}.
The UrbanSound8K dataset is a widely used collection of audio data designed for research in the field of urban sound recognition. UrbanSound8K consists of 8,732 audio clips, each lasting 4 seconds. These clips are extracted from longer field recordings and are labeled with specific sound classes. The dataset is annotated with 10 sound classes and we evaluate models on the test set with accuracy metric.

\noindent \textbf{LLVIP (LLVIP)} \cite{jia2021llvip}.
The LLVIP dataset features RGB and Thermal image pairs. Following the approach in ImageBind\cite{girdhar2023imagebind}, we use it for binary classification tasks, extracting pedestrian and random bounding boxes to create a balanced dataset of 15,809 boxes. Top-1 accuracy is the evaluation metric.

\noindent \textbf{N-ImageNet-1K (N-IN-1K)}~\cite{kim2021n}.
N-ImageNet-1K encompasses paired event data derived from the ImageNet-1K~\cite{deng2009imagenet} dataset. The evaluation focuses on assessing the event recognition capabilities of models within this dataset. Accuracy is employed as the metric for this evaluation.

\noindent \textbf{N-Caltech-101 (N-Cal)}~\cite{orchard2015converting}.
N-Caltech-101 includes paired event data related to the Caltech-101 dataset. This dataset is used for event recognition and both event-to-image and image-to-event retrieval tasks. Accuracy is the respective metric for recognition tasks.

\noindent \textbf{Touch-and-Go (TouchGo)}~\cite{yang2022touch}. 
Touch-and-Go is a dataset with paired visual and tactile data. This dataset is collected from probe objects in natural environments using tactile sensors, while simultaneously recording egocentric video. The metric for the recognition task is top-1 accuracy.


\begin{figure*}[t!]
    \centering
    \includegraphics[width=\linewidth]{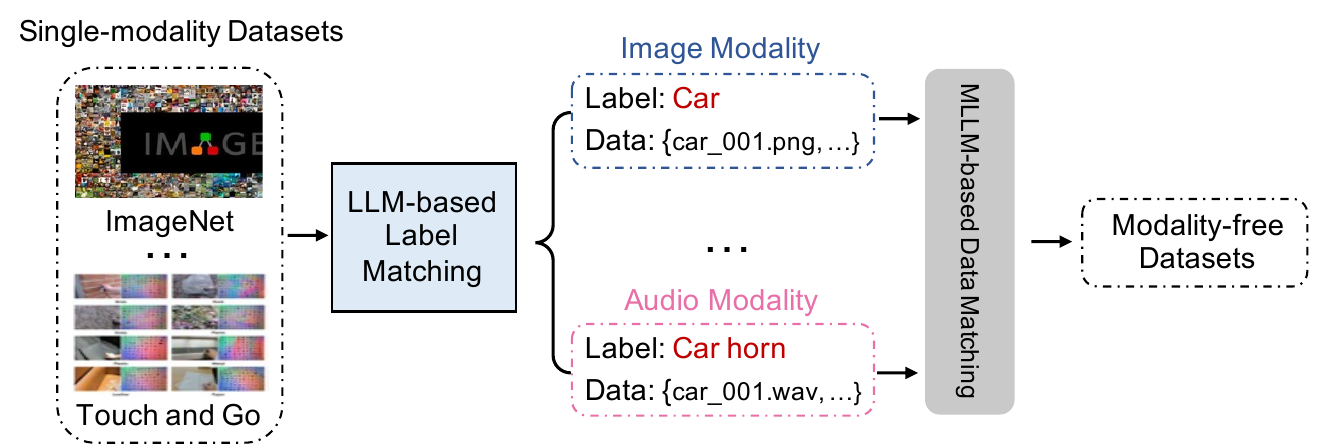}
    \caption{
    The Modality-free Dataset. We first match the data from various modalities in the label level via LLMs, and then we match these data in the data level via the similarities of the data descriptions generated by MLLMs.
    }
    \label{fig:dataset_collect}
\end{figure*}

\subsection{More Details of Modality-free Dataset}


\begin{figure}[t!]
\vspace{-10pt}
  \centering
  \includegraphics[width=\linewidth]{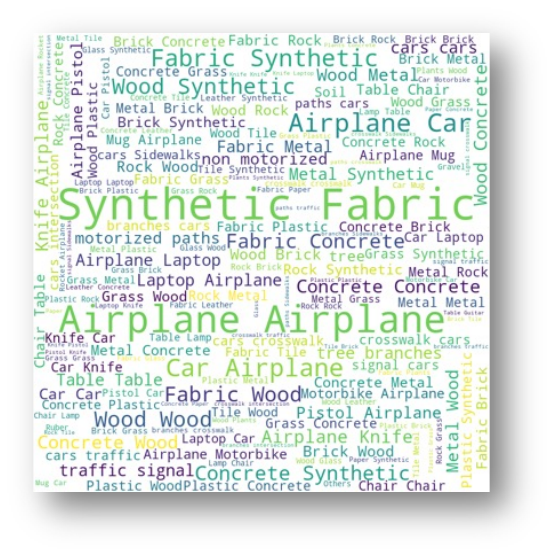}
    \vspace{-18pt}
    \caption{
   The word cloud of the modality-free dataset.
    }
    \vspace{-14pt}
    \label{fig:word_cloud}
\end{figure}

\begin{figure}[t!]
    \centering
    \includegraphics[width=\linewidth]{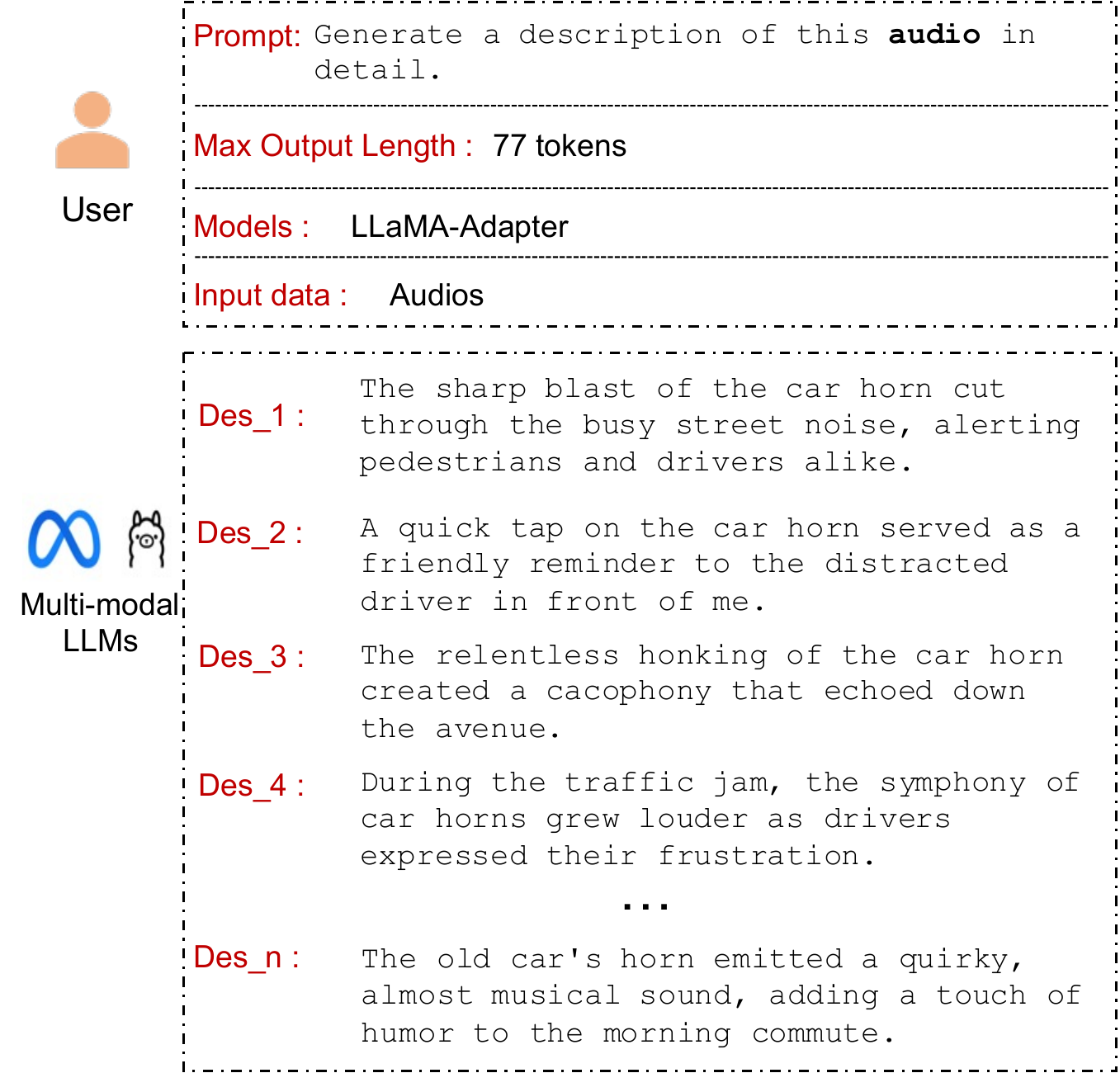}
    \caption{
    The details of description generation via MLLMs.
    }
    \label{fig:generation}
\end{figure}

We constructed a modality-free dataset from a range of source datasets, including ImageNet-1K, Caltech-101, ShapeNet, ESC-50, Urban-Sound-8K, LLVIP, N-ImageNet-1K, N-Caltech-101, and Touch-and-Go. As illustrated in Fig.~\ref{fig:dataset_collect}, we initially aligned the labels across seven modalities using the LLM (ChatGPT-4\cite{openai2023gpt4}) to ensure semantic correspondence between the categories. Subsequently, we utilized the MLLM (LLaMA-Adapter\cite{zhang2023llama}) to generate descriptive texts for various visual modalities, thereby facilitating sample-level matching through similarity calculations between text descriptions and visual data. The process of generating these descriptions is detailed in Fig.~\ref{fig:generation}.


Specifically, we derive audio and point cloud sets with matching semantics by analyzing and aligning the labels of both datasets using language modeling techniques. This alignment ensures that labels share identical meanings across modalities. Subsequently, as depicted in Fig.~\ref{fig:generation}, we employ the LLaMA-Adapter to generate descriptive texts for both the audio and point cloud data. We then establish correspondence between the audio files and point clouds based on the semantic content of their descriptions, calculating cosine similarity between the text embeddings of descriptions from the audio and point cloud datasets.

Lastly, we present the data distribution of each label as a word cloud in Fig.~\ref{fig:word_cloud}. It demonstrates the equalization of our dataset among categories.

\section{More Details of Our Proposed OmniBind}

We show more details of our proposed OmniBind in the demo code. We show the training code of knowledge-distilled alignment learning in code \textit{DemoKDAL.py}, and we show the code of our OmniBind model in code \textit{OmniBind.py}. We provide our project code and some cases of modality-free datasets in the \textit{supplementary material}.

\label{code_1}
\begin{lstlisting}[language=Python,title={DemoKDAL.py}]
def train(args, model, train_dataloader, device):
    real_batch_size = args.train_batch_size * args.gradient_steps
    t_total = train_dataloader.dataset.__len__() // 
    real_batch_size * args.num_train_epochs
    param_optimizer = list(model.named_parameters())
    optimizer_grouped_parameters = [
            {'params': [p for n, p in param_optimizer], 
            'weight_decay': 0.01}
        ]
    
    optimizer = AdamW(optimizer_grouped_parameters,
                lr=args.learning_rate, eps=args.adam_epsilon)
    scheduler = get_linear_schedule_with_warmup(
        optimizer, num_warmup_steps=args.warmup_steps, 
        num_training_steps=t_total)
    global_step = 0
    tr_loss = 0.0
    best_acc = 0.0
    writer = SummaryWriter(log_dir=(args.output_dir + '/tb_loss'))
    model.zero_grad()
    for epoch in range(int(args.num_train_epochs)):
        epoch_steps = len(train_dataloader)
        for step, batch in enumerate(train_dataloader):
            model.train()
            t_embeddings,text_embeddings,v_embeddings= 
            model(batch)
            logic = v_embeddings @ t_embeddings.t()
            labels = gen_label(logic, device)
            cl_loss = loss_fun(logic, labels)
            t_logic = t_embeddings @ text_embeddings.t()
            s_logic = v_embeddings @ t_embeddings.t()
            kl_loss = kldivloss(s_logic, t_logic)
            self_t_logic = t_embeddings @ t_embeddings.t()
            self_s_logic = v_embeddings @ v_embeddings.t()
            self_kl_loss = kldivloss(self_s_logic, self_t_logic)
            loss = (cl_loss+kl_loss*100+self_kl_loss*100)/3
            if args.gradient_accumulation_steps > 1:
                loss = loss / args.gradient_accumulation_steps
            loss.backward()
            tr_loss += loss.item()
            if (step + 1) % args.gradient_accumulation_steps == 0:
                global_step += 1
                torch.nn.utils.clip_grad_norm_(
                    model.parameters(), args.max_grad_norm)
                writer.add_scalar(tag="cl_loss", scalar_value= 
                tr_loss / global_step, 
                global_step=step+epoch_steps*epoch)
                optimizer.step()
                scheduler.step()
                model.zero_grad()
                if global_step % args.eval_steps == 0:
                    logger.info('Start eval!')
                    acc = evaluate()
                    logger.info('Dev acc: {0}'.format(acc))
                    if acc >= best_acc:
                        best_acc = acc
                        torch.save(
                            model.state_dict(), 
                            args.ckpt_dir
                        )
    writer.close()
    torch.save(model.state_dict(), args.ckpt_dir)

\end{lstlisting}

\label{code_3}
\begin{lstlisting}[language=Python,title={OmniBind.py}]
class OmniBind(nn.Module):
    def __init__(self, args):
        super(OmniBind, self).__init__()
        self.modality = args.modality
        self.backbone = UniBind.load(args.pre_train_weights)
        for param in self.backbone.parameters():
            param.requires_grad_(False)
        self.fuse_adapter = Adapter(args.dim, args.head_num)
    def forward(self, inputs):
        embeddings = self.backbone(inputs)
        fuse_embedding = self.fuse_adapter(embeddings).mean(dim=1)
        fuse_embedding /= fuse_embedding.norm(dim=-1, keepdim=True)
        return fuse_embedding
\end{lstlisting}

\begin{table}[ht!]
\renewcommand{\tabcolsep}{8pt}
\resizebox{\linewidth}{!}{
\begin{tabular}{llccc}
\toprule
\multicolumn{1}{c}{Modality} & \multicolumn{1}{c}{Dataset} & \multicolumn{1}{c}{batch size} & \multicolumn{1}{c}{lr} & \multicolumn{1}{c}{total epochs}\\ \midrule
Touch & Touch-and-Go~\cite{yang2022touch} & 512 & 5e-4 & 20 \\
Thermal & LLVIP~\cite{jia2021llvip} & 256 & 1e-3 & 20 \\
Event & N-Caltech-101~\cite{orchard2015converting} & 128 & 5e-4 & 20 \\
All & Modality-free & 512 & 1e-3 & 10 \\
\bottomrule
\end{tabular}}
\caption{The hyperparameters of experiments with the proposed OmniBind. }
\label{hypara}
\end{table}

\begin{table}[h!]
\renewcommand{\tabcolsep}{8pt}
\resizebox{\linewidth}{!}{
\begin{tabular}{l|cc|c} 
\toprule
Method & Combination 7 & Combination 8 &  Total \\ \midrule
UniBind~\cite{lyu2024unibind} & 33.60 & 80.90 & 48.37 \\ 
UniBind-linear~\cite{lyu2024unibind} & 87.20 & 90.40 & 87.48 \\ 
Ours & \textbf{93.40} & \textbf{99.24} & \textbf{92.42} \\ 
\rowcolor{gray!10}$\Delta $ & \textbf{+6.20} & \textbf{+8.84} & \textbf{+4.94} \\
\bottomrule
\end{tabular}}
\caption{The additional experiment results on the modality-free dataset. Combination 7: image, touch, and event; Combination 8: point cloud, audio, event, and text.}
\label{tab: compare_all}
\end{table}

\begin{table}[h!]
\renewcommand{\tabcolsep}{2pt}
\resizebox{\linewidth}{!}{
\begin{tabular}{l|ccc|c} 
\toprule
\multicolumn{1}{l|}{Method} & Three Modalities & Four Modalities & Five Modalities & Total\\ \midrule
UniBind~\cite{lyu2024unibind}+mean & 17.23 & 61.82 & 64.25 & 46.81 \\ 
UniBind~\cite{lyu2024unibind}+linear & 60.22 & 91.98 & 93.92 & 80.39 \\
Ours w/o Neg. Label & 70.74 & 90.68 & 93.27 & 83.20 \\
Ours & \textbf{78.20} & \textbf{93.36} & \textbf{94.75} & \textbf{86.94} \\ 
\rowcolor{gray!10}$\Delta $ & \textbf{+7.46} & \textbf{+2.68} & \textbf{+1.48} & \textbf{+3.74} \\ 
\bottomrule
\end{tabular}}
\caption{The ablation study for exploring the robustness of OmniBind.}
\label{tab: ab_noise}
\end{table}

\section{Training Setup}
Our experiments were done on 80GB A100 GPUs, and we detail the hyperparameters used for training for stages 1 and 2 reported in Tab.~\ref{hypara}.

\section{Additional Results}
We showcase more results with the combinations of three modalities and four modalities in Tab.~\ref{tab: compare_all}. For these additional combination inputs, our OmniBind achieves \textit{\textbf{an average performance gain of 4.94\%}}.


\section{Additional Ablation studies}


\begin{table*}[t!]
\renewcommand{\tabcolsep}{10pt}
\resizebox{\linewidth}{!}{
\begin{tabular}{l|cccccc} 
\toprule
Noise Modality&  \multicolumn{2}{|c}{Three Modalities} & \multicolumn{2}{c}{Four Modalities} & \multicolumn{2}{c}{Five Modalities}  \\ \cmidrule{2-7}
&  \multicolumn{1}{|c}{Combination 1}& Combination 2& Combination 3& Combination 4& Combination 5& \multicolumn{1}{c}{Combination 6}  \\ \midrule
Null & 96.78 & 90.40 & 89.60 & 98.65 & 92.75 & 98.48 \\ \midrule
text & 96.78 & 6.8 & 8.9 & 98.65 & 37.58 & 52.27 \\ 
\rowcolor{gray!10}$\Delta $ & - & -83.60 & -80.70 & - & -55.17 & -46.21 \\ \midrule
Event & 96.78 & 77.00 & 80.8 & 95.42 & 87.93 & 93.25 \\ 
\rowcolor{gray!10}$\Delta $ & - & -13.40 & -8.8 & -3.23 & -4.82 & -5.23 \\ \midrule
Touch & 96.78 & 75.40 & 79.01 & 98.65 & 85.90 & 98.48 \\ 
\rowcolor{gray!10}$\Delta $ & - & -15.00 & -10.59 & - & -6.85 & - \\
\bottomrule
\end{tabular}}
\caption{Effect of different noise modality on the performance. Combination 1: image, audio, and point cloud; Combination 2: text, event, and touch; Combination 3: text, event, touch, and image; Combination 4: image, event, audio, and point cloud; Combination 5: text, event, touch, image, and audio; Combination 6: image, point cloud, audio, event, and text}
\vspace{-12pt}
\label{ab: noise_m}
\end{table*}

To further analyze the robustness of our OmniBind framework, we conduct ablation studies by introducing noise into specific modalities to assess the model's robustness to noise across different modalities. As depicted in Tab.~\ref{ab: noise_m}, the performance significantly decreases when noise is added to the teacher modalities (text and image). In contrast, OmniBind exhibits a strong resilience to noise in the event and touch modalities. This observation suggests that the primary modality continues to play a dominant role in the interaction of multiple modalities.

\end{document}